# Gravity Optimizer: a Kinematic Approach on Optimization in Deep Learning

**Dariush Bahrami [1]; Sadegh Pouriyan Zadeh [2]**

[1] Faculty of New Sciences & Technologies, University of Tehran, Tehran, Iran

[2] Faculty of New Sciences & Technologies, University of Tehran, Tehran, Iran
Email: spouriyanz@ut.ac.ir

Corresponding Author; Email: dariush.bahrami@ut.ac.ir; Tel.: +98-910-192-8706

## Abstract

We introduce Gravity, another algorithm for gradient-based optimization. In this paper, we explain how our novel idea change parameters to reduce the deep learning model's loss. It has three intuitive hyper-parameters that the best values for them are proposed. Also, we propose an alternative to moving average. To compare the performance of the Gravity optimizer with two common optimizers, Adam and RMSProp, five standard datasets were trained on two VGGNet models with a batch size of 128 for 100 epochs. Gravity hyper-parameters did not need to be tuned for different models. As will be explained more in the paper, to investigate the direct impact of the optimizer itself on loss reduction no overfitting prevention technique was used. The obtained results show that the Gravity optimizer has more stable performance than Adam and RMSProp and gives greater values of validation accuracy for datasets with more output classes like CIFAR-100 (Fine).

Keywords:
Optimization Algorithm; deep learning; gradient descent; objective function

## 1. Introduction

The question of choosing an adequate optimizer for a deep learning problem is not answered yet. Instead, there are ways like empirical comparing [1–3] or benchmarking [4] which help to find better configurations for optimization.

The most common optimization techniques in deep learning are SGD (Stochastic Gradient Descent) [5], RMS Prop [6], and Adam [7]. Table 1 shows the most common standard optimization algorithms in chronicle order.

Table 1. Common standard optimizers in deep learning in chronicle order

| Year Published | Optimization technique |
|---|---|
| 1951 [5] | SGD |
| 1964 [8] | SGD with momentum |
| 2011 [9] | AdaGrad |
| 2012 [10] | AdaDelta |
| 2012 [6] | RMSProp |
| 2013 [11] | SGD with Nesterov momentum |
| 2015 [7] | Adam |
| 2015 [7] | AdaMax |
| 2016 [12] | Nadam |
| 2018 [13] | AMSGrad |

A lot of research has been done on optimizers in recent years which have introduced various optimizers [14–20]. Many studies have been done on them to compare their performance [2,21–23]. In algorithm 1, our proposed back-prop-based optimization method is given. The details of the Gravity optimization method with a novel kinematic approach will be given in section 2.

**Algorithm 1:** Here is the Gravity, our proposed optimization method with a kinematic approach. $\mathcal{N}$ is the normal distribution with a mean of $\mu$ and a standard deviation of $\sigma$. Also, $G$ is the gradient of the objective function, *J, w.r.t W*. The symbol $\oslash$ is the element-wise division (Hadamard division). This algorithm has three hyper-parameters whose recommended values are $l = 0.1$, $\alpha = 0.01$, $\beta = 0.9$. For easier implementation of the Gravity optimizer, its python implementation using TensorFlow's high-level API, Keras, is available in the Gravity GitHub repository.

**Require:** $l$: Learning Rate
**Require:** α: Govern initial Step size
**Require:** β: Moving Average Parameter ∈ [0,1]
**Require:** $t_{max}$: maximum number of update steps

for each weight matrix $W^i$:

$$\mu \leftarrow 0$$
$$\sigma \leftarrow \alpha/l$$
$$V_0^i \leftarrow \mathcal{N}(\mu, \sigma)$$
$$t \leftarrow 0$$
**while** $t < t_{max}$ :
$\quad t \leftarrow t + 1$
$\quad \hat{\beta} \leftarrow (\beta t + 1)/(t + 2)$
$\quad$ for each weight matrix $W^i$:
$\quad\quad G \leftarrow \partial J/\partial w$
$\quad\quad m \leftarrow 1 \oslash max(abs(G))$
$\quad\quad \zeta \leftarrow G \oslash (1 + (G \oslash m)^2)$
$\quad\quad V_t^i \leftarrow \hat{\beta} V_{t-1}^i + (1 - \hat{\beta})\zeta$
$\quad\quad W^i \leftarrow W^i - l V_t^i$

The rest of this article consists of the following sections. Section 2 describes the theory and mathematics of the Gravity optimizer. In the following, the behavior and effect of each hyper-parameter are explained and at the end of this section, the best-obtained hyper-parameters are suggested. Section 3 presents the tools used for the benchmark (including hardware, framework, and dataset) and the architecture chosen. Then the settings used in the optimizers (including hyper-parameters) are reported in detail. In Section 4, the obtained results from the training of each dataset on the selected architecture are reported in their subsections. At the end of each subsection, the performance of the Gravity optimizer is analyzed relative to the other two standard optimizers used. The final section provides a conclusion and what needs to be done in the future on the Gravity optimizer.

## 2 Gravity Optimizer Design

This section provides ideas, theories, and mathematics about the Gravity optimizer. Considering an inclined plane and using basic kinematic physics, an interesting analogy can be found between the gradient of the parameters in the deep learning models and the slope angle. In this analogy, the loss is equivalent to the height of the rolling ball. Fig. 1 shows the schematic of the idea.

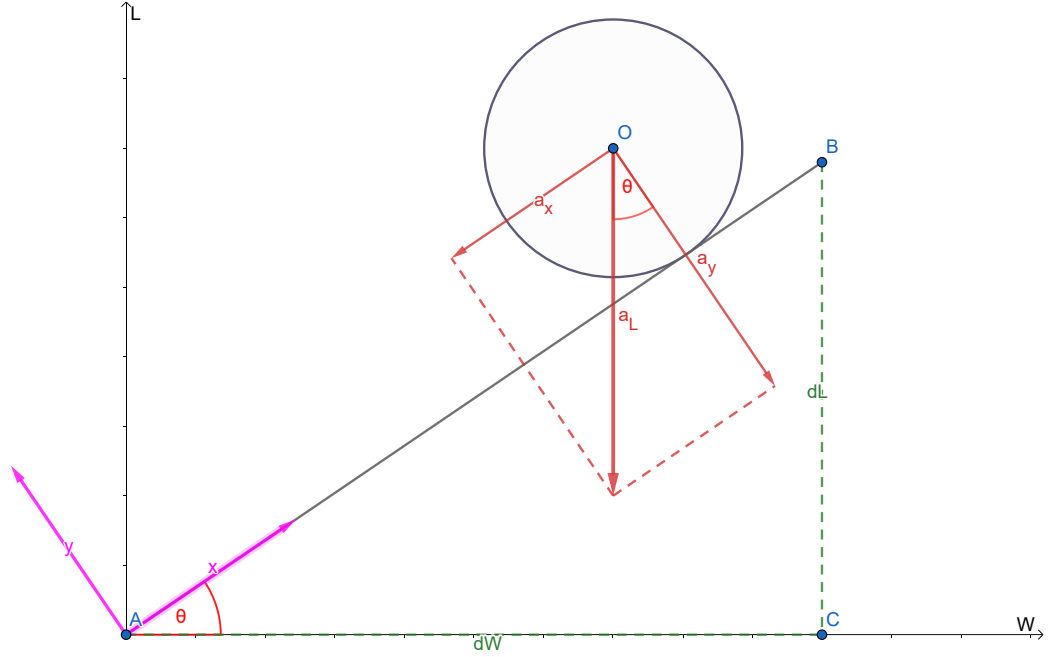

*Figure 1. The global coordinate system $<W, L>$ and the inclined plane's local coordinate system $<x, y>$*

For simplicity, the ball is treated as a point mass. The slope angle of the inclined plane, $\theta$, is obtained by the definition of the tangent:

$$tan(\theta) = \frac{dL}{dW} \quad \Rightarrow \quad \theta = tan^{-1}\left(\frac{dL}{dW}\right) \tag{1}$$

For the sake of brevity $g$ is defined as follow:

$$g = \frac{dL}{dW}$$

A universal acceleration in the L axis direction, $a_L$, is defined to imitate gravity from which the idea of the name of the Gravity optimizer came. The relationship between the inclined plane's local coordinate system, $<x, y>$, and the global coordinate system, $<W, L>$, can be obtained by using basic trigonometric relationships:

$$W = x \cdot cos\theta \quad \Rightarrow \quad \Delta W = \Delta x \cdot cos\theta \tag{2}$$

$$L = x \cdot sin\theta \quad \Rightarrow \quad \Delta L = \Delta x \cdot sin\theta \tag{3}$$

Then the relationship between acceleration in the mentioned coordinate systems can be written as:

$$a_x = -a_L . \sin\theta \tag{4}$$

Two parameters on the right side of Eq. 4, $a_L$ and $\theta$, does not change at each update step. Therefore position equation for constant acceleration [24] can be written as:

$$x = \frac{1}{2} . a_x . t^2 + v_{0_x} . t + x_0 \tag{5}$$

Assuming $v_0 = 0$ at each update step and simplifying give:

$$\Delta x = \frac{1}{2} . a_x . t^2 \tag{6}$$

Then by substituting Eq. 6 into Eq. 2:

$$\Delta W = \frac{1}{2} . a_x . t^2 \cos(\theta) \tag{7}$$

which, together with Eq. 4 gives:

$$\Delta W = -\frac{1}{2} . a_L . t^2 . \cos(\theta) . \sin(\theta) \tag{8}$$

Also, trigonometric equations of sine and cosine of $tan^{-1}$ [25] are:

$$\sin(tan^{-1}(x)) = \frac{x}{\sqrt{1+x^2}} \tag{9}$$

$$\cos(tan^{-1}(x)) = \frac{1}{\sqrt{1+x^2}} \tag{10}$$

Thus using Eq. 1, Eq. 8, Eq. 9, and Eq. 10, finally gives $\Delta W$ as:

$$\Delta W = -\frac{1}{2} . a_L . t^2 . \frac{g}{1+g^2} \tag{11}$$

Eq. 11 is the parameter-update equation. As can be seen, there are a lot of hyper-parameters in this equation that needs a lot of time to be tuned. Besides, they are not intuitive. In the following subsections, these hyper-parameters will be replaced with more familiar and common hyper-parameters.

2.1 Learning Rate

In deep learning models, it is common to work with more familiar hyper-parameters like learning rate. To have more common hyper-parameters, the learning rate in Eq. 11 is defined as follows:

$$l = \frac{1}{2} \cdot a_L \cdot t^2 \tag{12}$$

Substituting Eq. 12 into Eq. 11 gives:

$$\Delta W = \frac{-lg}{1+g^2} \tag{13}$$

To learn more about the above equation, consider back-prop based optimization methods as a function of the gradient. Each optimization method takes the gradient as input and gives a step in the output for updating parameters (weight, bias, etc.). For example, Gradient Descent (without any modification) is a linear function of the gradient:

$$\Delta W = -lg \tag{14}$$

which is plotted as follows:

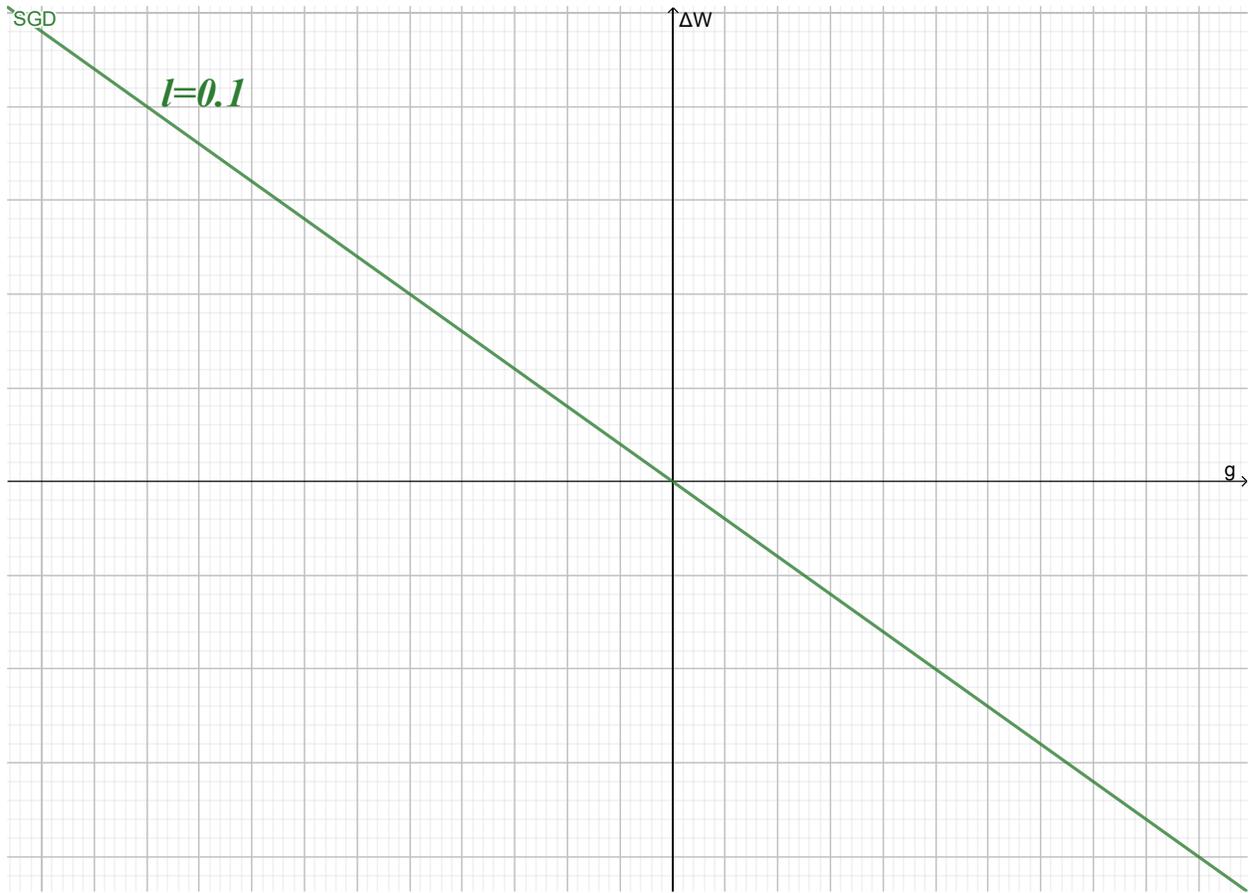

*Figure 2. Gradient Descent as a linear function of the gradient*

To compare the Gravity optimization method with Gradient Descent (GD), their functions are plotted in Fig. 3 for a given value of the learning rate:

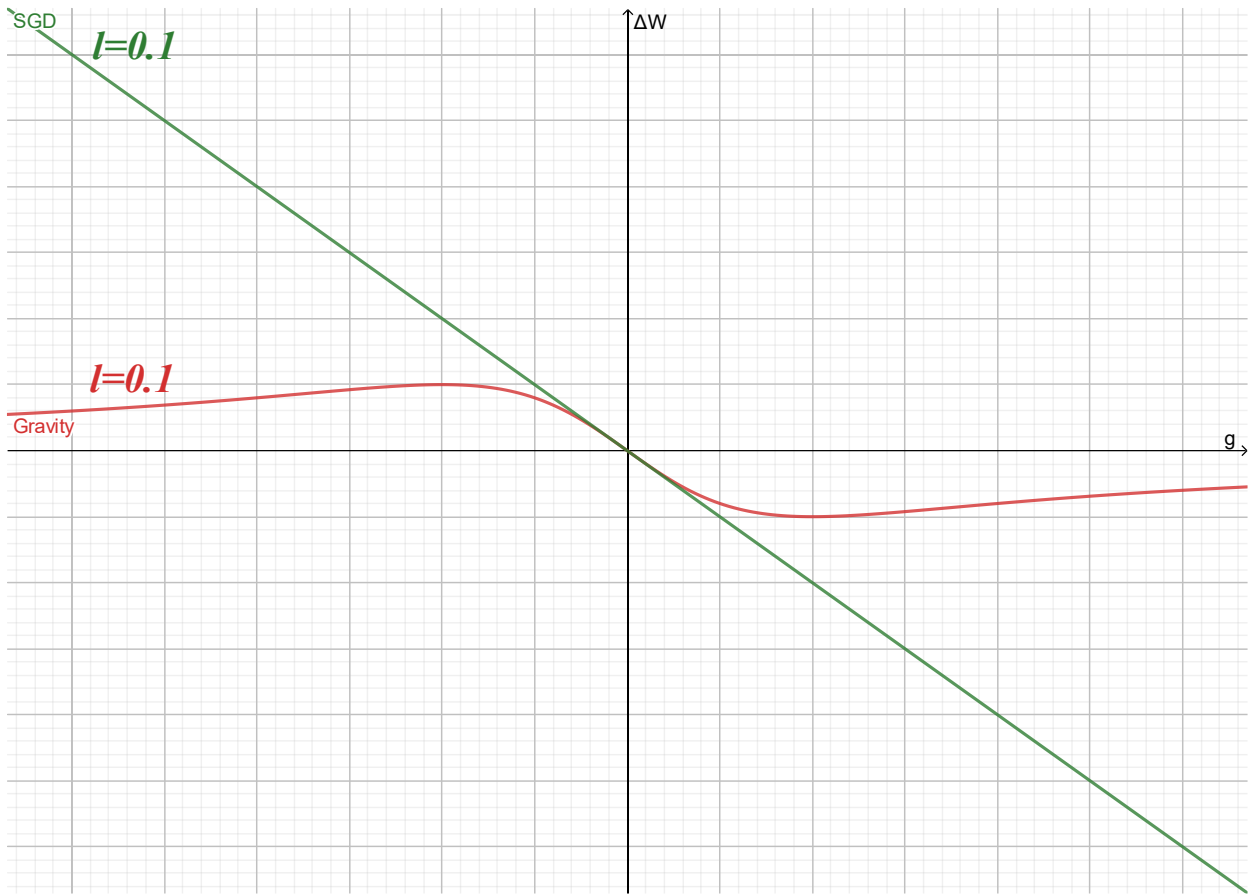

*Figure 3. Gravity and GD optimization functions for learning rate value of 0.1*

In Fig. 3, it is clear that for small gradient values, the Gravity behaves like GD. Mathematically: $\lim_{g \to 0} Gravity(g) = GD(g)$. To determine how small the gradient value must be for this similar behavior to occur, Fig. 4 is drawn for different values of the learning rate:

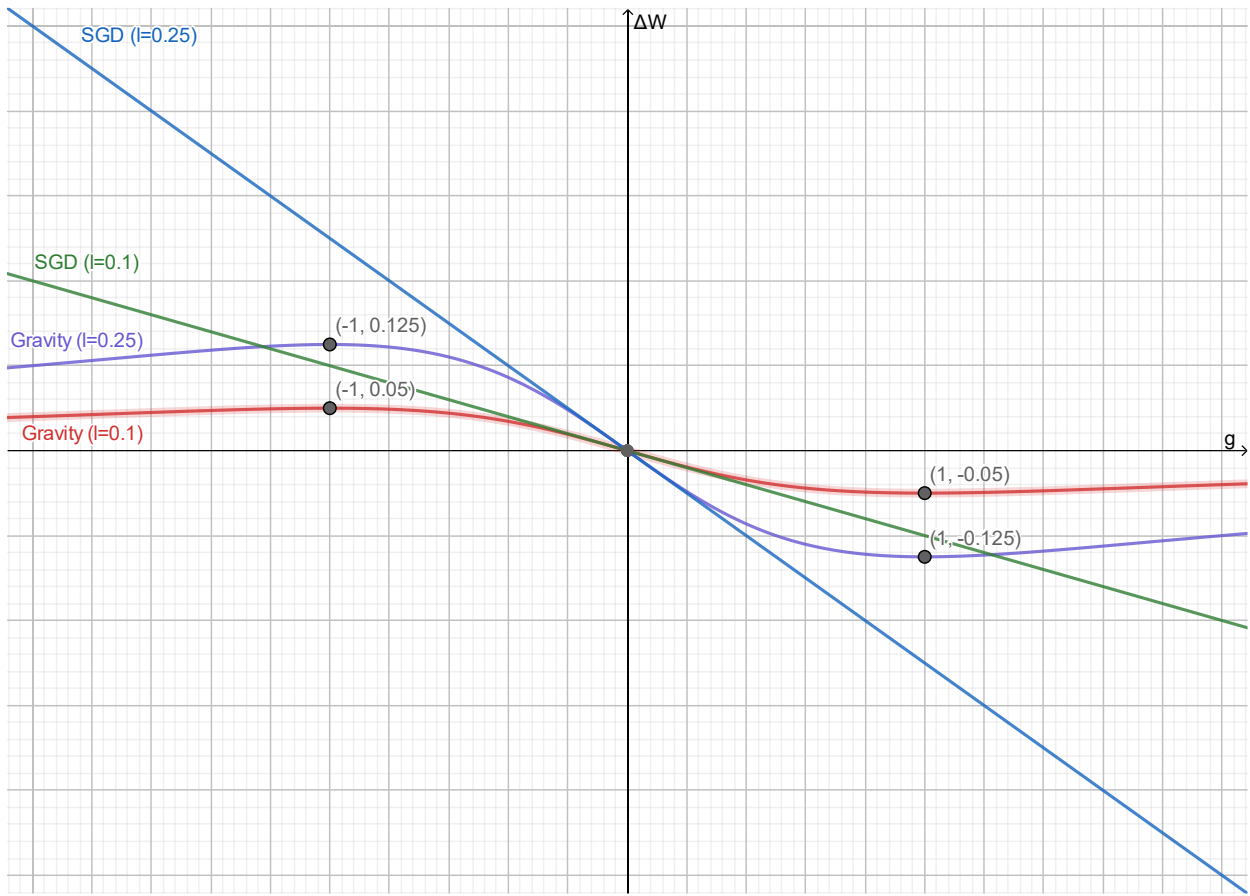

*Figure 4. Gravity and GD optimization functions for learning rate values of 0.1 and 0.25*

In Fig. 4, it can be seen that by changing the learning rate, the extremum of gravity which is the gradient at which the maximum step is taken (maximum output value of the optimization function) does not change. Therefore, whatever learning rate is selected, the maximum step for updating the parameter occurs at g = 1, which corresponds to 45 degrees.

2.2 Max-Step Grad

So far, we have transferred the two parameters we encountered in the Gravity optimization function ($a_L$ and $t$) to learning rate hyper-parameter. The values of these two parameters are the same for all the weights of the weight matrix i.e. they have the same values for two different weights. But to gain more control over the optimization function, we need to have a parameter to change the slope angle θ.

We know that on inclined planes, we can watch a falling ball at a slower speed by reducing the slope angle (like a slow-motion video taken with a high-speed camera).

In contrast by increasing the slope angle, θ, the falling time will be reduced and everything happens quickly.

Therefore, with more control over this angle, we can reach the minimum loss with more speed and less divergence. It can be done by tweaking gradient with a coefficient called Max-Step Grad, $m$:

$$tan(θ) = \frac{1}{m}\frac{dL}{dW} \quad \Rightarrow \quad θ = tan^{-1}\left(\frac{1}{m}\frac{dL}{dW}\right) \tag{15}$$

which $m > 0$. Plugging new θ from Eq. 15 into Eq. 8 and using Eq. 9 and Eq. 10 to simplify, new $\Delta W$ with more control is obtained as follows:

$$\Delta W = -\frac{1}{2}.a_L.t^2.\frac{\frac{g}{m}}{1+\left(\frac{g}{m}\right)^2} \tag{16}$$

The learning rate for this more controlled optimization function is defined as:

$$l = \frac{1}{2m}.a_L.t^2 \tag{17}$$

Substituting Eq. 17 into Eq. 16 finally gives:

$$\Delta W = \frac{-lg}{1+\left(\frac{g}{m}\right)^2} \tag{18}$$

For a deeper understanding of the Max-Step Grad hyper-parameter, Fig. 5 shows the effect of different values on the optimization function (with the same learning rate value of 0.1).

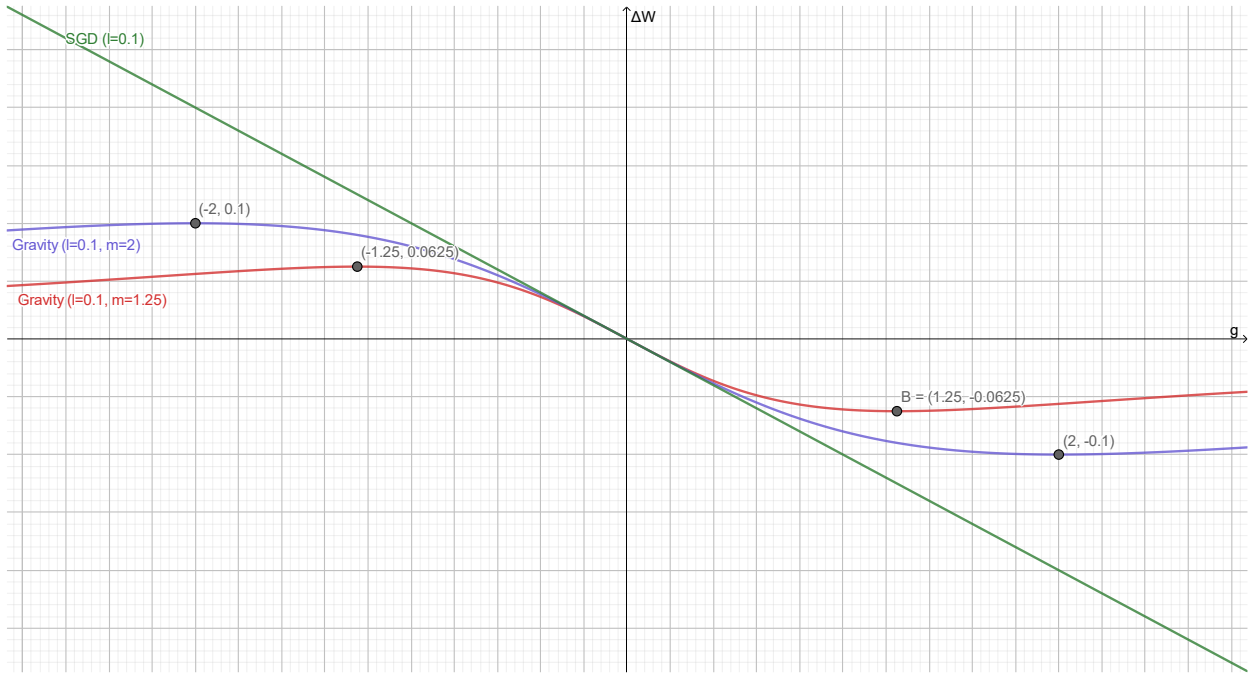

*Figure 5. Gravity and GD optimization functions for Max-Step Grad values of 1.25 and 2 (learning rate = 0.1)*

As mentioned at the beginning and shown in Fig. 5, m is the gradient in which the biggest steps are taken to change the weight. Mathematically it can be said:

$$\text{if} \quad f(g) = \Delta W = \frac{-lg}{1+\left(\frac{g}{m}\right)^2} \quad \text{and} \quad f'(g) = \frac{\partial f}{\partial g} \quad \text{then} \quad f'(m) = 0.$$

Now the maximum step ($\Delta W_{max}$) for given Max-Step Grad ($m$) and learning rate ($l$) can be obtained as follows:

$$\Delta W_{max} = \frac{lm}{2} \tag{19}$$

In summary, Max-Step Grad has two effects: the first is on the linear region of the curve and the second is on the maximum step value. A greater value of Max-Step Grad results in a larger section of the linear region as well as bigger steps for weights with higher gradients. In other words, a wider range of gradients is treated linearly by increasing m, and also weights having larger gradient values take larger steps.

2.2.1 a little about Gradient Descent divergence

The reason for the divergence in the Gradient Descent for large learning rates is weights with large gradients. Given the fact that the computers have limited capacity for storing large values for very steep slopes of the cost function, an infinite amount of $\Delta W$ is possible in the Gradient Descent optimization method. Here is a scenario that leads to divergence:

1. Consider a weight with a large gradient.
2. This weight takes a bigger step relative to others (linearly proportional to their gradient ratio)
3. After applying the optimizer, the weight goes too far and now has a larger gradient which leads to another big step
4. steps 1 to 3 will happen again to diverge.

As can be seen in [Fig. 6](), Harrington [26] explains that more intuitively:

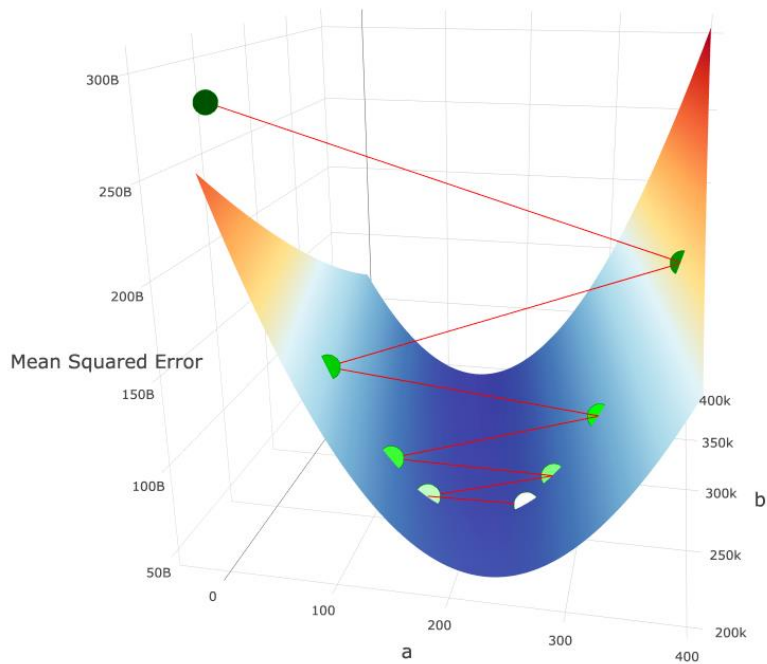

*Figure 6. The illustration of divergence in the Gradient Descent optimization method*

1. We start at the white point in the "valley" and calculate the gradient at that point.
2. Then we multiply the learning rate by the gradient and move along this vector to the new point (the slightly greenish point to the left of the white point). Because the learning rate was so high, combined with the magnitude of the gradient, we "jumped over" the local minimum.

3. Then again we calculate the gradient at point 2 and make the next move. Again, we jump over the local minimum. The gradient at point 2 is even greater than the gradient at point 1. So in the following steps, we again jump over the local minimum to diverge.
4. Due to the convex, valley-like curve of our objective function, as we continue to jump from side to side, the gradient at each jump grows higher. The error increases quadratically with each "jump", and the algorithm diverges to infinite error.

2.3 Choosing m (Max-Step Grad)

It is necessary to limit $\Delta W$ for weights with large values of $g$ to avoid divergence. Given the fact that gradients are constantly changing during training, it is clear that the value of $m$ cannot be determined in advance. We suggest that the value of m be selected based on the current gradient matrix. Knowing that a gradient matrix with larger gradients must have a smaller $m$ to avoid divergence, we suggest selecting m as follows:

$$m = \frac{1}{max(abs(G))} \tag{20}$$

which G is the gradient matrix. In other words, to select m with a geometrical interpretation of Eq. 20, the following steps can be done:

1. find the largest gradient or in other words steepest $\theta$
2. calculate the complementary angle correspond to it: $\alpha = \frac{\pi}{2} - \theta$
3. and choose $m = tan(\alpha)$

2.4 Moving Average

Most of the common optimizers in deep learning like SGD with momentum, Adam, and RMSProp use moving average to stabilize loss reduction. We tested the exponential moving average on the Gravity optimizer and the results were promising. The main issue before applying moving average was an initial delay in loss reduction even though the optimizer does its job very well after some initial epochs with no loss reduction. To solve this issue, the gradient term, $\zeta$ , is defined as follow:

$$\zeta = \frac{g}{1+\left(\frac{g}{m}\right)^2} \tag{21}$$

also, the velocity, $V$, is defined as follow:

$$V_t = \beta V_{t-1} + (1-\beta)\zeta \tag{22}$$

In Eq. 22, $\beta$ is a positive real number which $0 < \beta < 1$. Also, $V_t$ is the value of $V$ in current update step (mini-batch) and $V_{t-1}$ is $V$ at the previous update step. The value of $V$ is initialized with 0 at $t = 0$. By these definitions, new update rule based on $V$ is as follow:

$$\Delta W_t = -lV_t \tag{23}$$

The effect of $\beta$ on loss reduction is shown in Fig. 7 by comparison of different values of $\beta$.

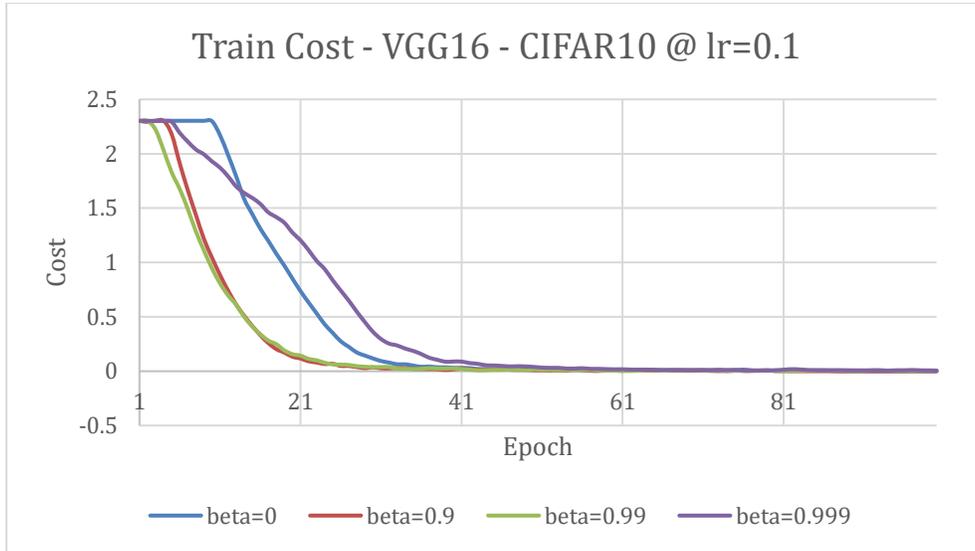

*Figure 7. The effect of different values of β on loss reduction*

After performing some experiments that included changing the value of β, we finally found that the optimal value for β was 0.9 in most cases. Although tuning may be required in some special cases. While moving average helped the Gravity with initial speed, there were still some delays. To solve this problem, we propose an alternative value for β.

For any specific value of β, the average is taken over a certain number of previous data. The number of these data can be calculated by Eq. 24 [27].

$$\text{number of averaged data} \approx \frac{1}{1-\beta} \tag{24}$$

In the first epochs, there is not enough data to be averaged. Also, the value of Eq. 22 will be very small at the beginning of the training because $V_0 = 0$. There is a solution known as bias correction which modifies $V_t$ as follow:

$$V_t = \frac{\beta V_{t-1} + (1-\beta)\zeta}{1-\beta^t} \tag{25}$$

The denominator of Eq. 25 approaches to 1 by increasing the value of $t$ and this makes that Eq. 22 and Eq. 25 output almost the same result. But at the beginning of the training, the value of $1 - \beta^t$ is small, so dividing Eq. 22 to this small value increases the total outcome.

We tried to use bias correction but in large values of β (closer to 1) we encountered overflow. Therefore we propose an alternative to β to solve the problem of Eq. 22 at initial steps with a different approach:

$$\hat{\beta} = \frac{\beta t + 1}{t+2} \tag{26}$$

Value of $\hat{\beta}$ in Eq. 26 at large values of $t$ will tend to β but at smaller values of $t$ will correct the value of β. The derivation of Eq. 26 is described as follows:

Choosing any value for β, the average will be over almost $1/(1 - \beta)$ number of previous data. We define $\hat{\beta}$ a variable alternative of β which increases over time and tends to 1 so the average will be always over all of the previous data. For averaging over all of the previous data in each step, the amount of data that will be averaged over should be $t + 2$; because at $t = 0$, there are two data available, $V_0$ and $V_1$. This can be written as follow:

$$\frac{1}{1-\hat{\beta}} = t + 2 \quad \Rightarrow \quad \hat{\beta} = \frac{t+1}{t+2}$$

At $t = 0$ the value of $\hat{\beta}$ is 0.5 which will average over two data ($V_0$ and $V_1$). By increasing the value of $t$, the value of $\hat{\beta}$ tends to 1 which will average over all of the previous data. Table 2 shows the value of $\hat{\beta}$ for different values of $t$:

Table 2. values of $\hat{\beta}$ for different values of t and $\beta = 1$

| t | $\hat{\beta}$ | Available Data | Averaged Data |
|---|---|---|---|
| 0 | 0.5 | 2 | 2 |
| 1 | 0.6667 | 3 | 3 |
| 2 | 0.75 | 4 | 4 |
| 3 | 0.8 | 5 | 5 |
| 4 | 0.8333 | 6 | 6 |
| 5 | 0.8571 | 7 | 7 |
| 6 | 0.875 | 8 | 8 |
| 7 | 0.8889 | 9 | 9 |
| 8 | 0.9 | 10 | 10 |
| 9 | 0.9090 | 11 | 11 |
| 10 | 0.9166 | 12 | 12 |
| 11 | 0.9230 | 13 | 13 |
| 12 | 0.9285 | 14 | 14 |
| 13 | 0.9333 | 15 | 15 |
| 14 | 0.9375 | 16 | 16 |
| 15 | 0.9411 | 17 | 17 |

As can be seen in Table 2, the number of data on which the average is done in each step is equal to the total number of available data. For averaging over $1/(1-\beta)$ number of previous data, the above equation is modified as:

$$\hat{\beta} = \frac{\beta t + 1}{t + 2}$$

which is the proposed alternative to β, Eq. 26. The value of $\hat{\beta}$ tends to β by increasing the value of $t$. The behavior of $\hat{\beta}$ for different values of β is shown in Fig. 8.

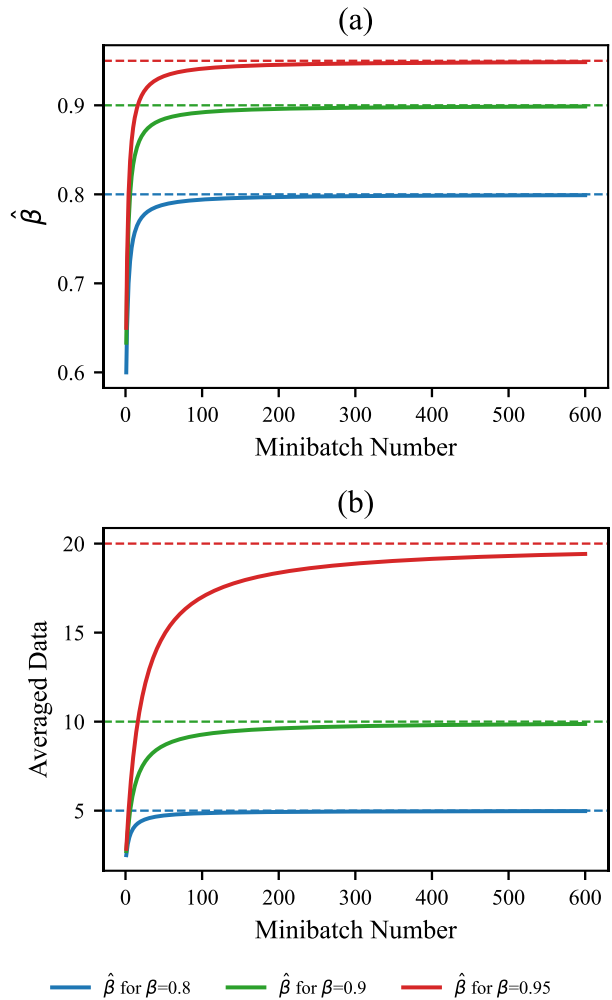

*Figure 8. (a) The behavior of $\hat{\beta}$ for different values of $\beta$ (b) The number of previous data on which average is taken*

In addition to modifying β in Eq. 22, we came out with another solution for increasing the speed of the Gravity optimizer at early steps by using non-zero initial $V$. Instead of zero, $V$ was initialized with normally distributed numbers with a mean (μ) of zero and a standard deviation of:

$$\sigma = \frac{\alpha}{l} \tag{27}$$

Take a closer look at the first update step:

$$\Delta W_0 = -l(0.5 V_0 + 0.5 \zeta_0) \tag{28}$$

Eq. 28 can be seen as two separate parts. The first part is $\Delta W_{V_0}$ which is due to initial $V$ and the second part is $\Delta W_{\zeta_0}$ which is due to the gradient term. For the second part, because it is about the gradient term and is determined by various parameters, nothing can be done. But the first part, $\Delta W_{V_0}$, can be tweaked to reach faster initial loss reduction speed. This term is defined as follows:

$$\Delta W_{V_0} = -0.5 l V_0 \tag{29}$$

If $V_0$ is initialized to zero, the total control of optimization will be at the gradient part of Eq. 28, $\Delta W_{\zeta_0}$, which is out of control.

It is shown in Fig. 9 that for a normally distributed set of random numbers, 68% of values are less than the standard deviation ($\sigma$), 95% are less than $2\sigma$, and 99.7% are less than $3\sigma$. Therefore by choosing $\sigma$, the range of initial steps for most of the different parameters (weights and biases) will be defined.

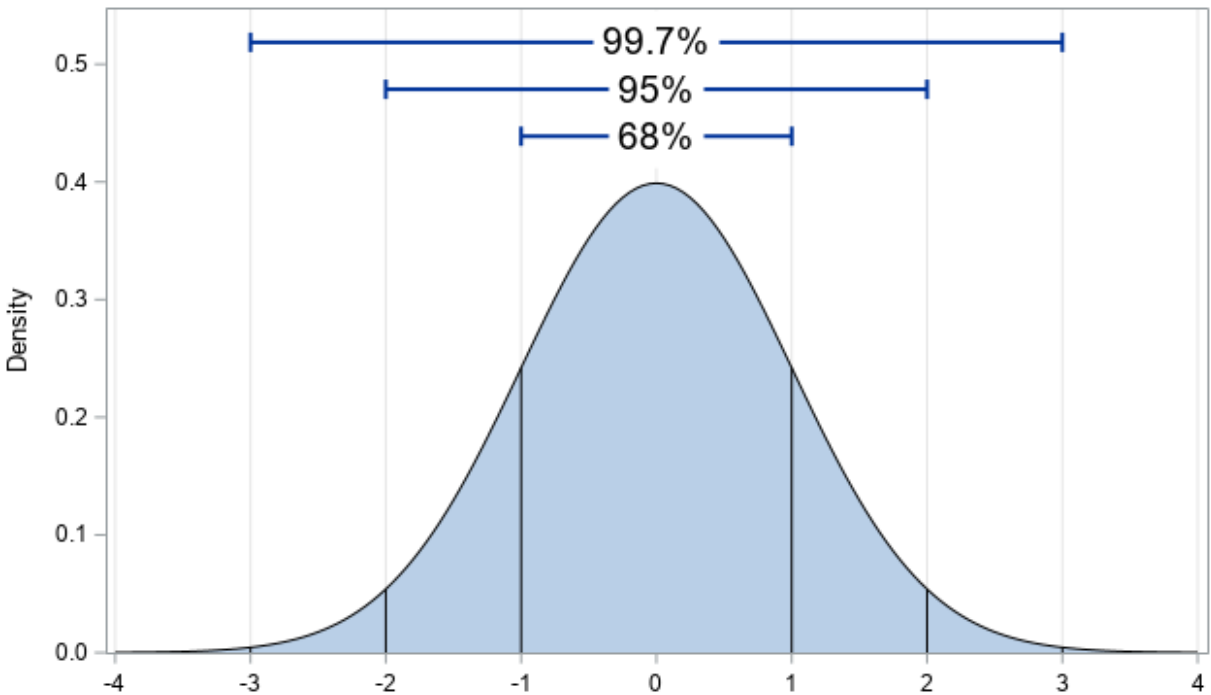

*Figure 9. the 68 – 95 - 99.7 rule for the normal distribution*

By selecting each $\sigma$, 95% of the parameters will be as follows:

$$|\Delta W_{V_0}| < l\sigma$$

The right side of the above inequality is defined as a new hyper-parameter, called alpha, $\alpha$, to give more control over the value of initial steps:

$$\alpha = l\,\sigma \tag{30}$$

Eq. 30 is a different form of Eq. 27. Experiments with different values of alpha α showed that α = 0.01 is satisfying for most models.

## 3. Benchmark Configuration

In this section, the Gravity optimizer is compared with two other standard optimizers shown in Table 1, Adam and RMSProp. In the following subsections, first, the specifications of the hardware used for the training are given. Then the framework used to implement the model, the datasets used for training, and finally, the architectures chosen based on hardware specifications are introduced. If you want to skip reading the details, a summary of datasets, models/architectures, and optimizers used is given in Table 3, Table 4, and Table 5 respectively. In the next section, the results obtained from training are reported.

3.1 Hardware and Framework

Google Colaboratory [28] (`colab.research.google.com`) is used as hardware because it was difficult for us to afford GPU for training deep neural network models and testing our ideas. Also, by using Google Colab and TensorFlow together we were given the chance to use the Tensor Processing Unit, TPU, Google's custom-developed technology to accelerate machine learning workloads. Using Google Colab showed us that free tools Google made available to everyone can help people put their ideas into action.

TensorFlow's high-level API, Keras, is used as the framework to implement the models in the Python programming language. The python implementation code can be found in the Gravity optimizer GitHub repository.

3.2 Datasets

We used the following standard datasets to evaluate the performance of the Gravity optimizer: MNIST, Fashion-MNIST, CIFAR-10, CIFAR-100 (Coarse), and CIFAR-100 (Fine). The MNIST database of handwritten digits is a subset of a larger set available from NIST. The images of digits have been size-normalized and centered in a fixed-size image [29]. The Fashion-MNIST is a dataset containing images of 10 classes. The 10 different classes are T-shirt, Trouser, Pullover, Dress, Coat, Sandal, Shirt, Sneaker, Bag, and Ankle boot [30]. CIFAR-10 is a subset of the 80 million tiny images dataset in 10 classes. The 10 different classes represent airplanes, cars, birds, cats, deer, dogs, frogs, horses, ships, and trucks [31]. CIFAR-100 is just like the CIFAR-10, except it has 100 classes containing 600 images each. The 100 classes in the CIFAR-100 are grouped into 20 superclasses. Each image comes with a "fine" label (the class to which it belongs) and a "coarse" label (the superclass to which it belongs). We trained the "fine" and "coarse" datasets separately because they have two distinct labels that classify two different types of classification; "coarse" is more general and "fine" is more specific[31]. Table 3 summarizes the detailed information of the datasets used.

Table 3. detailed information of datasets used for benchmark

| Dataset | Train # | Test # | Class # | Shape | Image per class |
|---|---|---|---|---|---|
| MNIST [29] | 60 K | 10 K | 10 | 28x28x1 | 6 K |
| Fashion-MNIST [30] | 60 K | 10 K | 10 | 28x28x1 | 6 K |
| CIFAR-10 [31] | 50 K | 10 K | 10 | 32x32x3 | 5 K |
| CIFAR-100 (coarse) [31] | 50 K | 10 K | 20 | 32x32x3 | 2.5 K |
| CIFAR-100 (fine) [31] | 50 K | 10 K | 100 | 32x32x3 | 500 |

3.3 Architecture (models and hyper-parameters)

The VGG Network architecture (VGG16 and VGG19) was used with the exact specifications reported in their paper [32]. VGG16 has about 34M and VGG19 has about 39M parameters for classification of 10 classes and the input shape of 32x32x3 (detailed number of parameters in each layer is shown in Table 4). Although architectures such as ResNet50 [33] and EfficientNet [34] have 23M and 4M parameters respectively (for 10 output classes and the input shape of 32x32x3) and they are as easy to implement as VGGNet in Keras, they showed so much slower training speed than VGGNet in Google Colab. The model summary of all of these architectures are available in the Gravity optimizer GitHub repository

Here, to investigate the direct impact of the optimizer itself on loss reduction, no kinds of overfitting prevention techniques were used; such as learning rate decay [35], dropout [36], early stopping [37], batch normalization [38], and regularization [39]. None of these techniques are used in the VGG Network architectures, so it makes it a good choice.

Table 4. VGG16 and VGG19 model summary used in the paper

| VGG16 | | | VGG19 | | |
|---|---|---|---|---|---|
| **Layer Type** | **Output Size** | **Parameters#** | **Layer Type** | **Output Size** | **Parameters#** |
| **Convolution Part** | | | **Convolution Part** | | |
| Input Layer | 32, 32, 3 | 0 | Input Layer | 32, 32, 3 | 0 |
| Conv2D | 32, 32, 64 | 1,792 | Conv2D | 32, 32, 64 | 1,792 |
| Conv2D | 32, 32, 64 | 36,928 | Conv2D | 32, 32, 64 | 36,928 |
| MaxPooling2D | 16, 16, 64 | 0 | MaxPooling2D | 16, 16, 64 | 0 |
| Conv2D | 16, 16, 128 | 73,856 | Conv2D | 16, 16, 128 | 73,856 |
| Conv2D | 16, 16, 128 | 147,584 | Conv2D | 16, 16, 128 | 147,584 |
| MaxPooling2D | 8, 8, 128 | 0 | MaxPooling2D | 8, 8, 128 | 0 |
| Conv2D | 8, 8, 256 | 295,168 | Conv2D | 8, 8, 256 | 295,168 |
| Conv2D | 8, 8, 256 | 590,080 | Conv2D | 8, 8, 256 | 590,080 |
| Conv2D | 8, 8, 256 | 590,080 | Conv2D | 8, 8, 256 | 590,080 |
| MaxPooling2D | 4, 4, 256 | 0 | Conv2D | 8, 8, 256 | 590,080 |
| Conv2D | 4, 4, 512 | 1,180,160 | MaxPooling2D | 4, 4, 256 | 0 |
| Conv2D | 4, 4, 512 | 2,359,808 | Conv2D | 4, 4, 512 | 1,180,160 |
| Conv2D | 4, 4, 512 | 2,359,808 | Conv2D | 4, 4, 512 | 2,359,808 |
| MaxPooling2D | 2, 2, 512 | 0 | Conv2D | 4, 4, 512 | 2,359,808 |
| Conv2D | 2, 2, 512 | 2,359,808 | Conv2D | 4, 4, 512 | 2,359,808 |
| Conv2D | 2, 2, 512 | 2,359,808 | MaxPooling2D | 2, 2, 512 | 0 |
| Conv2D | 2, 2, 512 | 2,359,808 | Conv2D | 2, 2, 512 | 2,359,808 |
| MaxPooling2D | 1, 1, 512 | 0 | Conv2D | 2, 2, 512 | 2,359,808 |
| **Dense Part** | | | Conv2D | 2, 2, 512 | 2,359,808 |
| Flatten | 512 | 0 | Conv2D | 2, 2, 512 | 2,359,808 |
| Dense | 4096 | 2,101,248 | MaxPooling2D | 1, 1, 512 | 0 |
| Dense | 4096 | 16,781,312 | **Dense Part** | | |
| Dense | 10 | 40,970 | Flatten | 512 | 0 |
| | | | Dense | 4096 | 2,101,248 |
| | | | Dense | 4096 | 16,781,312 |
| | | | Dense | 10 | 40,970 |
| **Total Parameters = 33,638,218** | | | **Total Parameters = 38,947,914** | | |

Finally, we monitor loss and accuracy changes for training and validation datasets for a constant number of epochs (100 epochs) to compare the performance of the Gravity optimizer with two common standard optimizers listed in Table 1 (Adam and RMSProp). Table 4 summarizes the models used in this paper.

The remarkable thing about the Gravity optimizer is that there was no need to tune hyper-parameters to get better results. The same values were chosen in all runs. In Section 2, the details of designing the hyper-parameters were discussed. The recommended value for the Gravity optimizer hyper-parameters is:

learning rate ($l$) = 0.1 , Alpha ($\alpha$) = 0.01 , Beta ($\beta$) = 0.9.

These values are set as default for the Gravity optimizer in python implementation and have been used for all training.

The activation function for all layers except the last layer is the ReLU function. It is defined as f(x) = max (0, x). As we know, the ReLU activation function was first used by Fukushima but not given any particular name [40]. Also Nair & Hinton's paper [41] spurred the recent interest in using the ReLU function in neural networks, and it is the source of the modern nomenclature "Rectified Linear Unit".

In the TensorFlow documentation, it is recommended not to use the Softmax function for multi-class classification[42]. instead, it is recommended to give logits (numeric output of the last linear layer of a multi-class classification neural network) directly to the cost function. Thus in the last layer, instead of using the Softmax function, classification is done by using sparse categorical cross-entropy class in Keras and turning "from logits" attribute to True.

Table 5 shows a detailed summary of the learning rate values used in runs. For Adam optimizer, learning rate decay was set off ($decay = 0$), $\beta_1, \beta_2$, and $\epsilon$ were set to 0.9, 0.999, and $1.0\,e-7$ respectively. For RMSProp, learning rate decay, momentum, and centered were set off ($decay = momentum = centered = 0$), $rho$ and $epsilon$ were set to $9.0\,e-01$ and $1.0\,e-07$ respectively.

Table 5. summary of learning rates used for benchmark

|  | VGG16 | | VGG19 | |
| --- | --- | --- | --- | --- |
|  | **Adam** | **RMSProp** | **Adam** | **RMSProp** |
| **MNIST** | 2.50e-04 | 1.00e-04 | 2.50e-04 | 2.50e-05 |
| **Fashion-MNIST** | 2.50e-04 | 5.00e-05 | 1.00e-05 | 5.00e-05 |
| **CIFAR-10** | 1.00e-04 | 5.00e-05 | 1.00e-04 | 2.50e-05 |
| **CIFAR-100 (coarse)** | 1.00e-04 | 2.50e-04 | 7.50e-05 | 1.00e-04 |
| **CIFAR-100 (fine)** | 1.00e-04 | 1.00e-04 | 5.00e-05 | 1.00e-04 |

To summarize, the training of five standard datasets mentioned in [subsection 3.2](#) was done on VGGNet architectures (VGG16 and VGG19) using the Gravity optimizer, RMSProp, and Adam using ReLU activation function with a batch size of 128 and for 100 epochs.

## 4. Results

In this section, the results obtained from the training of selected datasets on two VGGNet architectures using Gravity, Adam, and RMSProp optimization techniques are reported. Detailed information on hardware, framework, model/architecture, datasets, and hyper-parameters used is given in the previous section.

4.1 MNIST Results

In this subsection, results from training the MNIST dataset on VGG16 and VGG19 architectures without using any overfitting prevention techniques are reported. In [subsection 3.3](#) we discussed the reason for not using overfitting prevention techniques.

4.1.1 MNIST Results on VGG16

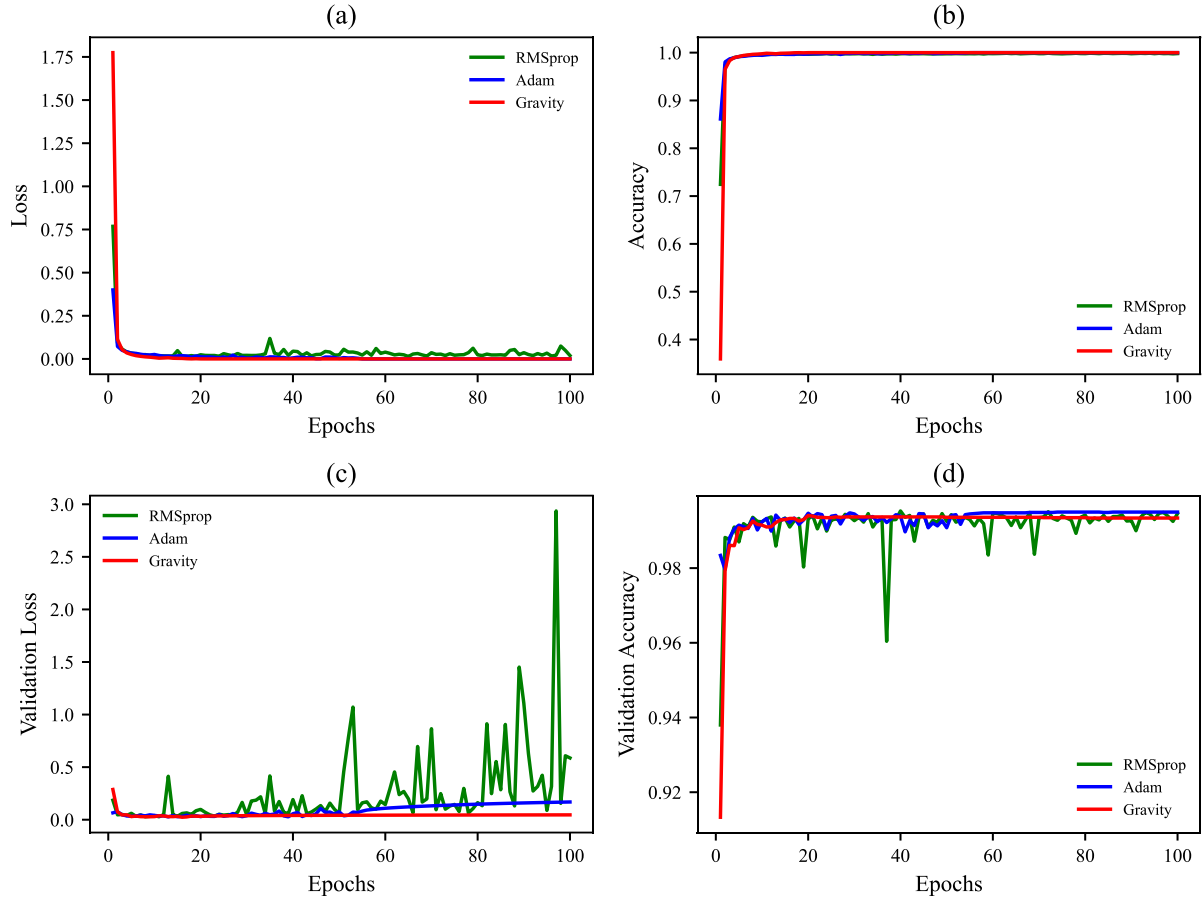

*Figure 10. Training MNIST on VGG16. learning rates used are shown in Table 5 (a) training loss vs. epochs (b) training accuracy vs. epochs (c) test loss vs. epochs (d) test accuracy vs. epochs*

Fig. 10 shows that the Gravity optimizer behaves as steadily as Adam although Adam shows more validation accuracy and Gravity has lower validation loss. The behavior of the RMSProp optimizer shows more oscillation in validation loss.

### 4.1.2 MNIST Results on VGG19

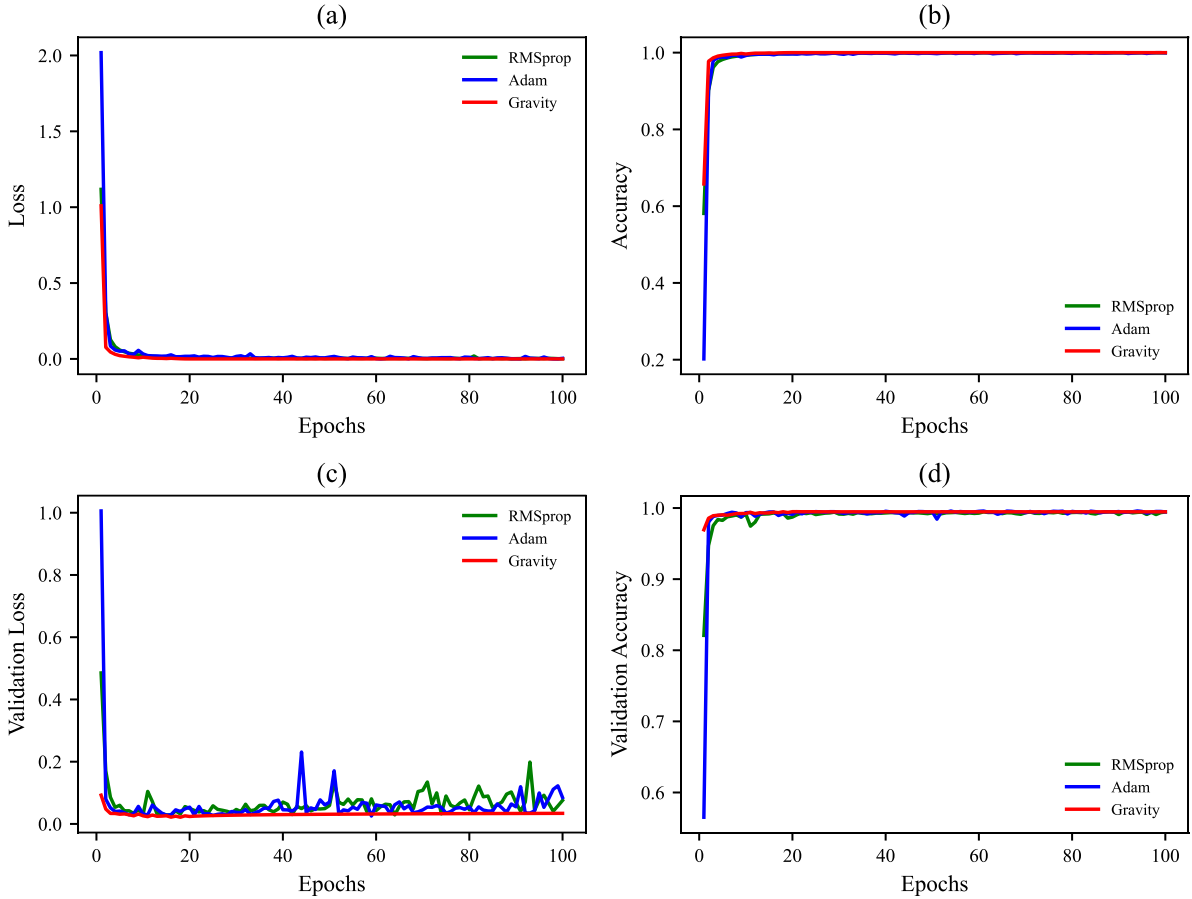

*Figure 11 Training MNIST on VGG19. learning rates used are shown in Table 5 (a) training loss vs. epochs (b) training accuracy vs. epochs (c) test loss vs. epochs (d) test accuracy vs. epochs*

As the network layers deepen from 16 to 19 layers, RMSProp oscillates less in validation loss and validation accuracy, Adam oscillates more in validation loss and oscillates less in validation accuracy. Also, Gravity has the least loss in validation loss and shows more accuracy in validation accuracy.

4.2 Fashion-MNIST

In this subsection, results from training the Fashion-MNIST dataset on VGG16 and VGG19 architectures are reported.

4.2.1 Fashion-MNIST on VGG16

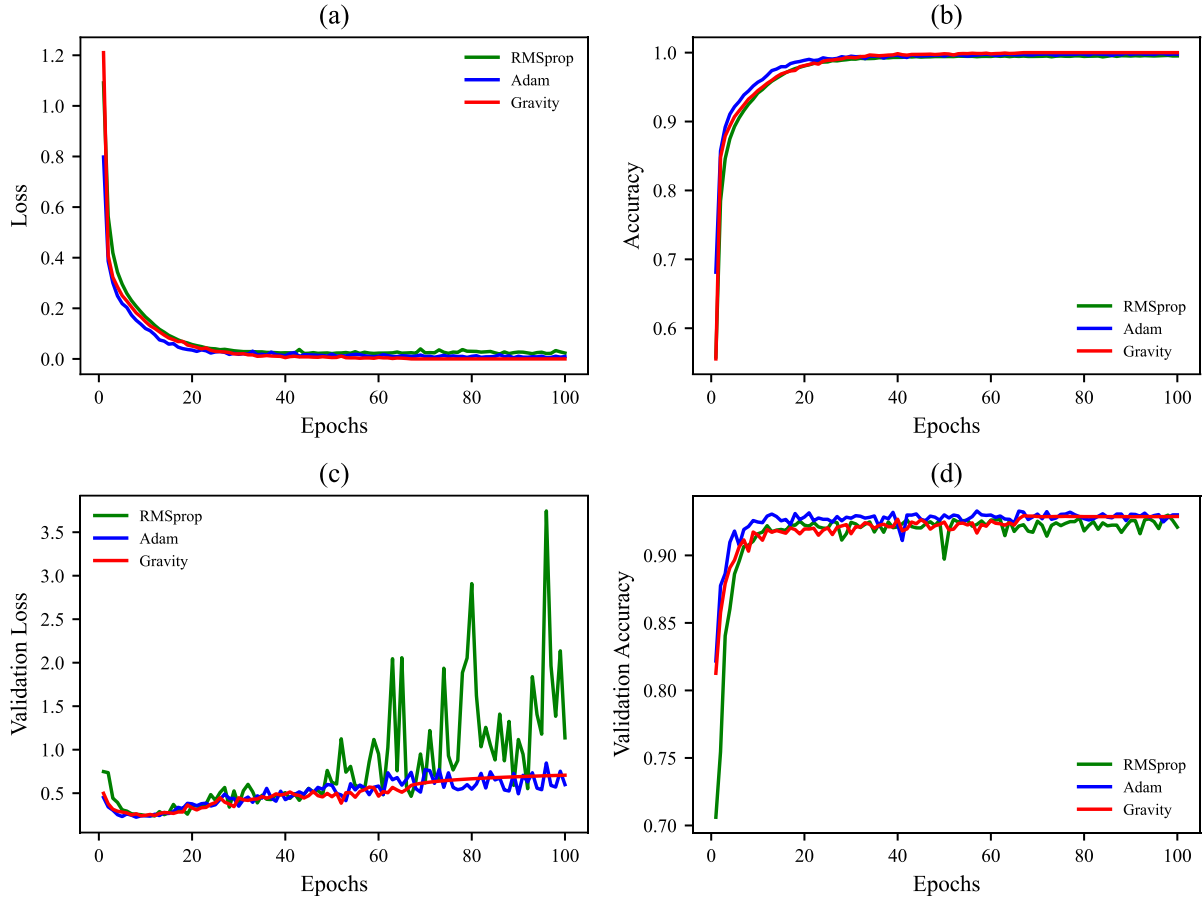

*Figure 12 Training Fashion-MNIST on VGG16. learning rates used are shown in Table 5 (a) training loss vs. epochs (b) training accuracy vs. epochs (c) test loss vs. epochs (d) test accuracy vs. epochs*

It can be seen in Fig. 12 that the gravity optimizer shows a more stable performance than Adam. There is also a lot of oscillation again in RMSProp for validation loss. Stability in the behavior of the gravity optimizer can be seen from epoch 60 onwards.

4.2.2 Fashion-MNIST on VGG19

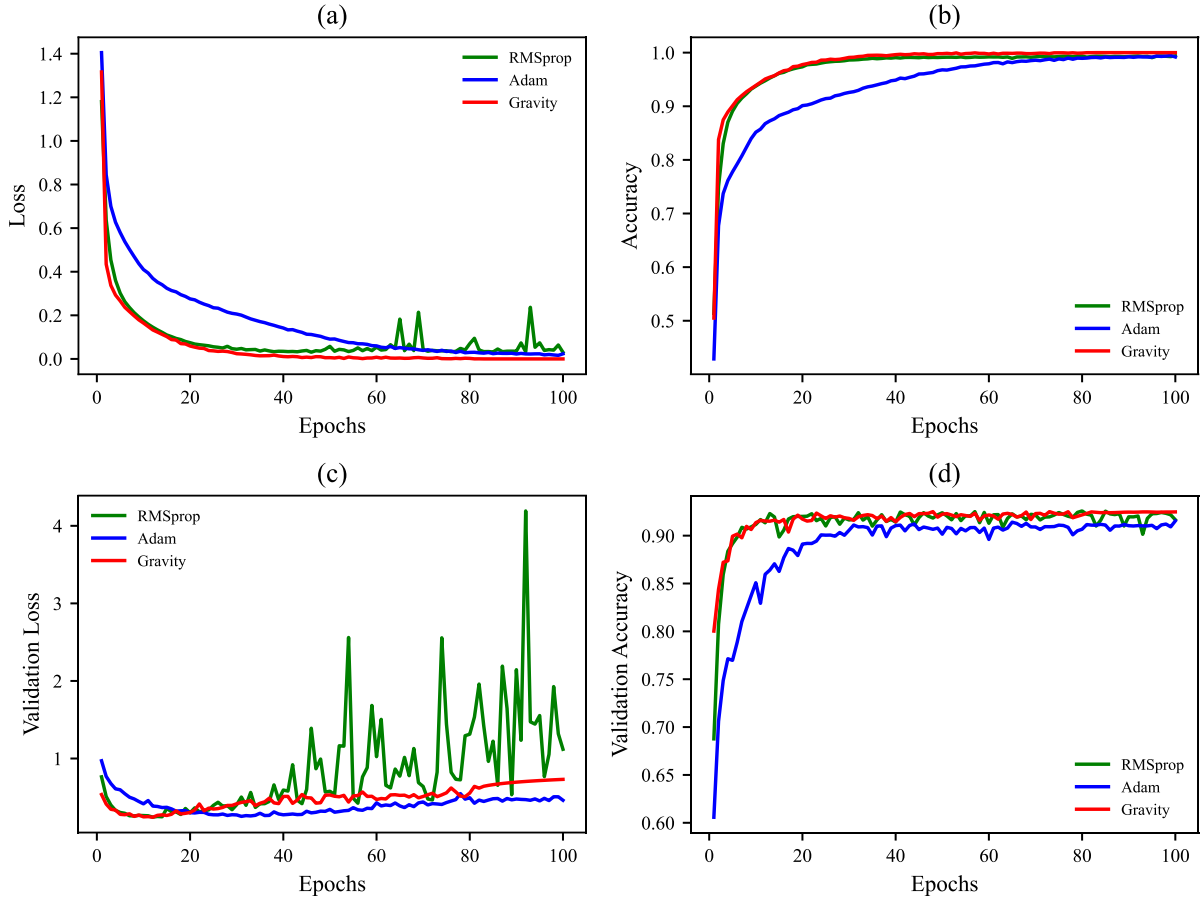

*Figure 13 Training Fashion-MNIST on VGG19. learning rates used are shown in Table 5 (a) training loss vs. epochs (b) training accuracy vs. epochs (c) test loss vs. epochs (d) test accuracy vs. epochs*

Here, as the depth of the layers increased from 16 to 19, a great change in Adam's behavior was observed; slower speed in reducing loss and increasing accuracy. Although Adam has reduced validation loss better than the other two optimizers, it did not show as much validation accuracy as them. A stable behavior in validation loss and validation accuracy of Gravity can be seen again from epoch 80 onwards. Unlike the reduction of the oscillatory behavior of the RMSProp by increasing the depth of the model on the training of the MNIST dataset, its oscillatory behavior did not decrease by deepening the model on the training of the Fashion-MNIST dataset.

4.3 CIFAR-10

In this subsection, results from training the CIFAR-10 dataset on VGG16 and VGG19 architectures are reported.

## 4.3.1 CIFAR-10 on VGG16

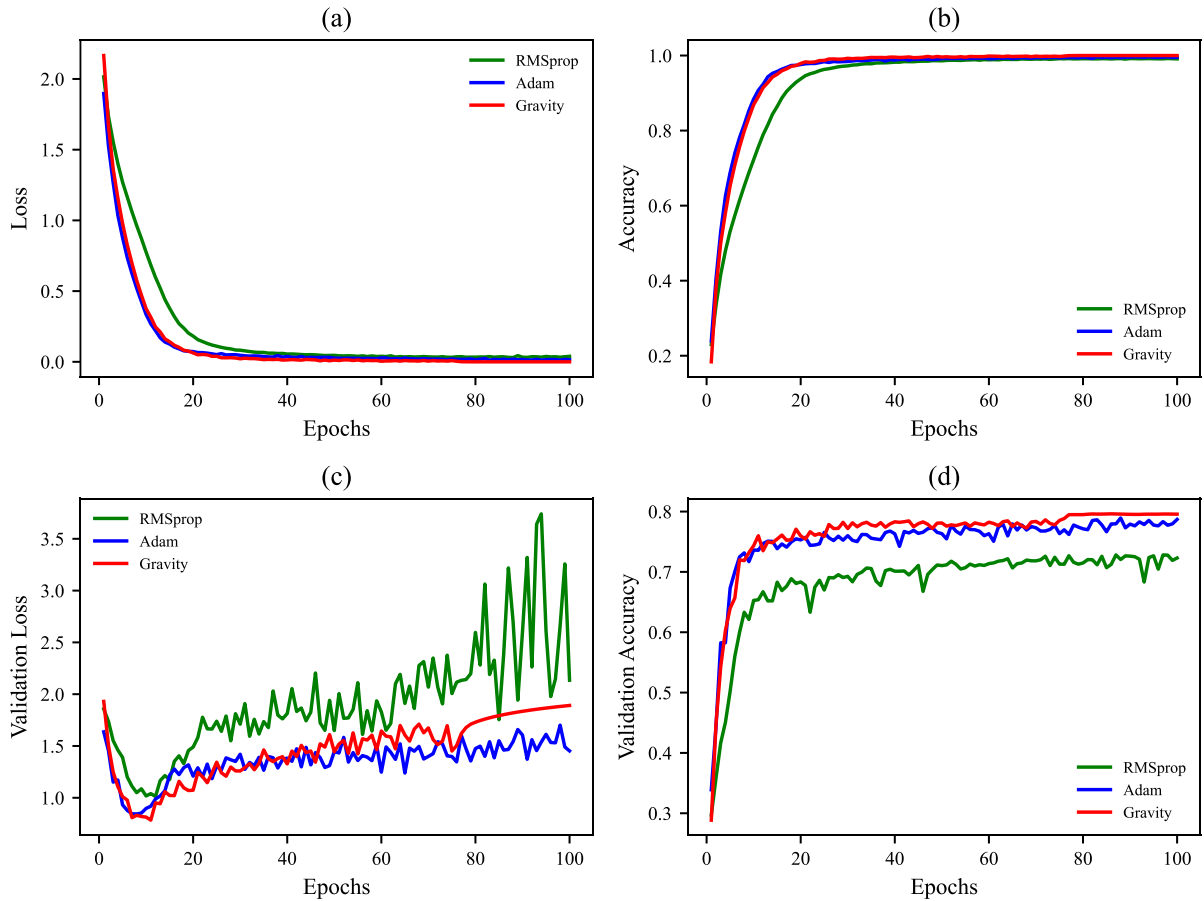

*Figure 14 Training CIFAR-10 on VGG16. learning rates used are shown in Table 5 (a) training loss vs. epochs (b) training accuracy vs. epochs (c) test loss vs. epochs (d) test accuracy vs. epochs*

Given the challenge of the CIFAR-10 dataset, all three optimizers have shown oscillation in dealing with the test dataset. But all have shown good results in reducing train loss and increasing train accuracy. Adam did a better job in reducing validation loss. Results of the Gravity in validation accuracy shows its better performance in generalization than the others. Also, the stable behavior of the Gravity is observed again from epoch 80 onwards.

## 4.3.1 CIFAR-10 on VGG19

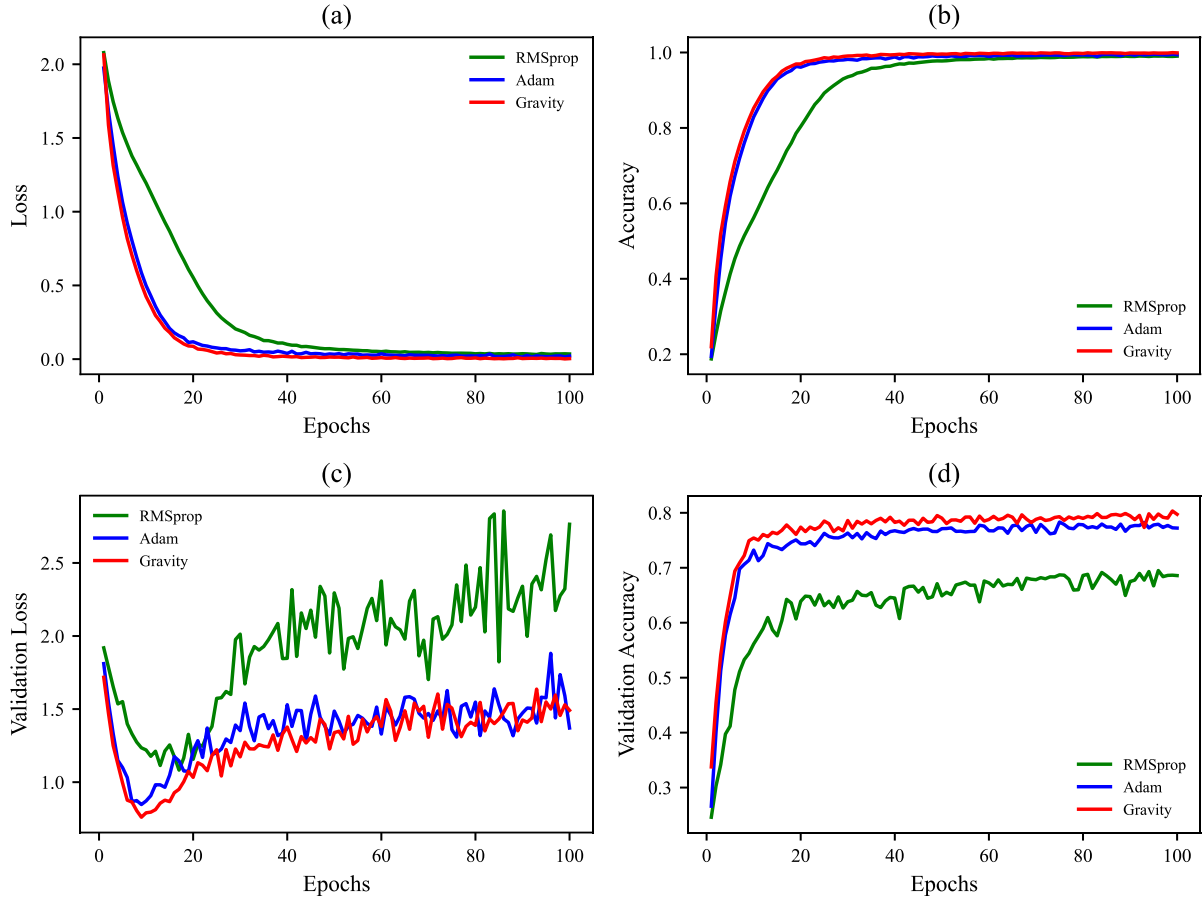

*Figure 15 Training CIFAR-10 on VGG19. learning rates used are shown in Table 5 (a) training loss vs. epochs (b) training accuracy vs. epochs (c) test loss vs. epochs (d) test accuracy vs. epochs*

Although RMSProp showed a similar result in the training dataset as Adam and Gravity for 100 epochs, it has shown a delay in reducing loss and increasing accuracy. It has also shown its oscillating behavior in reducing validation loss and also has not shown good results in increasing validation accuracy relative to Adam and Gravity.

Adam and Gravity have shown similar behavior in dealing with the CIFAR-10 dataset, although Gravity has been more successful in increasing validation accuracy like its performance in training CIFAR-10 on VGG16.

## 4.4 CIFAR-100 (Coarse)

In this subsection, results from training the CIFAR-100 (Coarse) dataset on VGG16 and VGG19 architectures are reported.

### 4.4.1 CIFAR-100 (Coarse) on VGG16

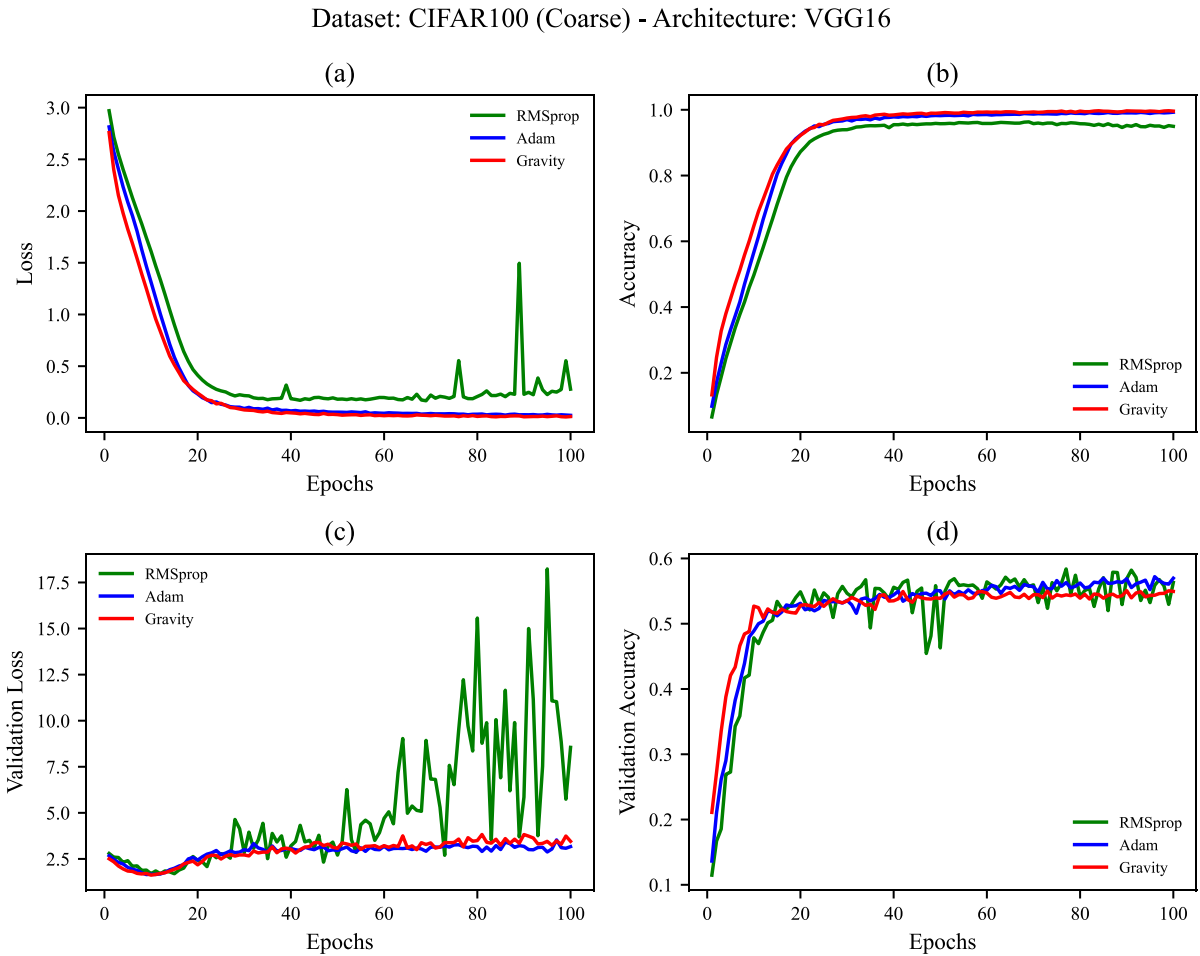

Figure 16 Training CIFAR-100 (Coarse) on VGG16. learning rates used are shown in Table 5 (a) training loss vs. epochs (b) training accuracy vs. epochs (c) test loss vs. epochs (d) test accuracy vs. epochs

Adam and Gravity behave similarly again, like in the training of the CIFAR-10 dataset, except that Adam reached less validation loss and more validation accuracy. RMSProp behaves worse than the other two optimizers and it still shows oscillatory behavior especially in reducing validation loss.

### 4.4.2 CIFAR-100 (Coarse) on VGG19

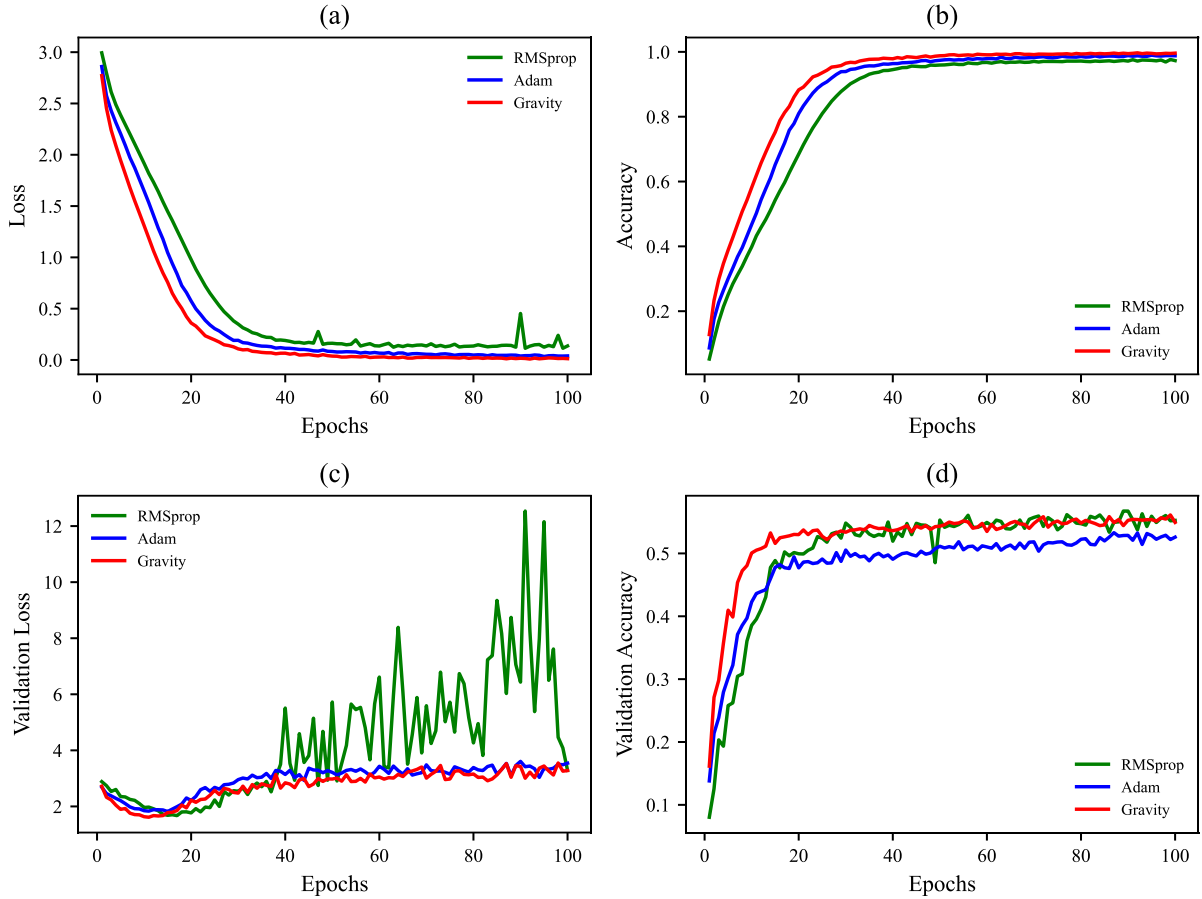

*Figure 17 Training CIFAR-100 (Coarse) on VGG19. learning rates used are shown in Table 5 (a) training loss vs. epochs (b) training accuracy vs. epochs (c) test loss vs. epochs (d) test accuracy vs. epochs*

In this data, which is the first dataset with more output class numbers than 10, Gravity showed better behavior than two others, reaching less train loss and more train accuracy. There is a big difference between Gravity and Adam in validation accuracy, a factor for comparing optimizers generalization. Again RMSProp oscillatory behavior is observed in validation loss. Also, it could not reduce train loss as much as the other two optimizers in all epochs.

4.5 CIFAR-100 (Fine)

In this subsection, results from training the CIFAR-100 (Fine) dataset on VGG16 and VGG19 architectures are reported.

4.5.1 CIFAR-100 (Fine) on VGG16

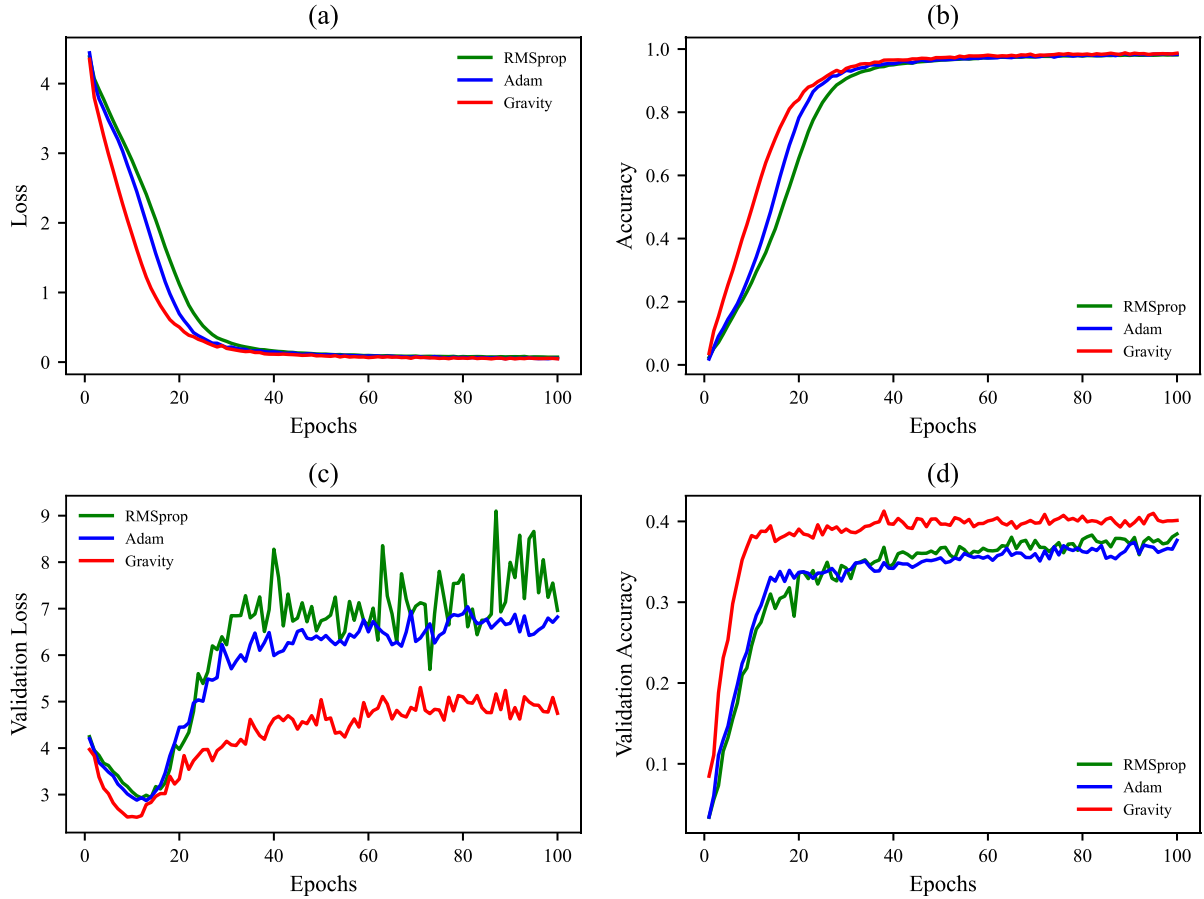

*Figure 18 Training CIFAR-100 (Fine) on VGG16. learning rates used are shown in Table 5 (a) training loss vs. epochs (b) training accuracy vs. epochs (c) test loss vs. epochs (d) test accuracy vs. epochs*

As shown in Table 3, CIFAR-100 has 100 classes and has created many challenges for optimizers and even deep networks. As shown in Fig. 18, although all three optimizers in the training dataset have reached similar values for loss and accuracy, the Gravity reduces training loss and increases training accuracy faster. The main difference in the behavior of all three optimizers can be seen in the test dataset, in which the gravity optimizer behaved better than the other two standard and widely used optimizers, Adam and RMSProp; less value for validation loss, more value for validation accuracy, and the faster speed in increasing validation accuracy.

4.5.2 CIFAR-100 (Fine) on VGG19

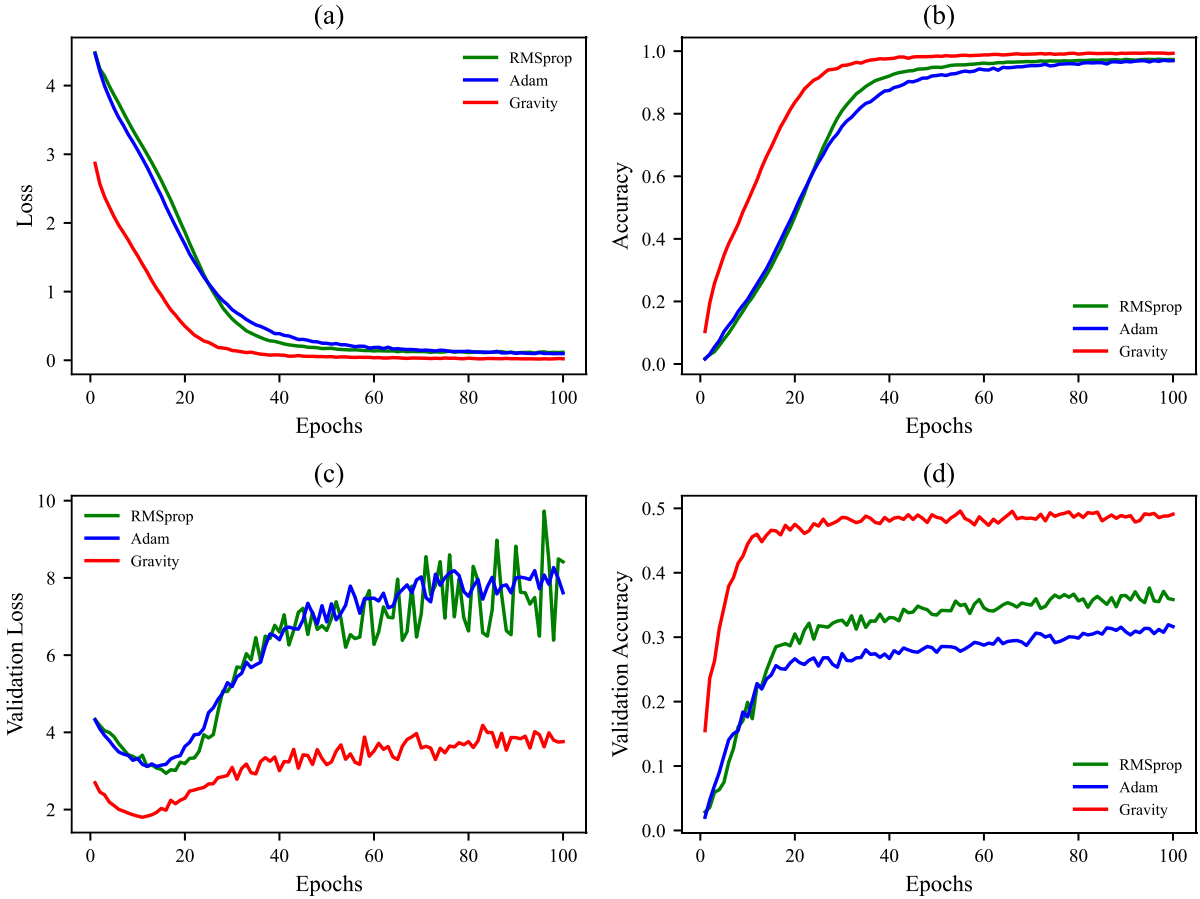

*Figure 19 Training CIFAR-100 (Fine) on VGG19. learning rates used are shown in Table 5 (a) training loss vs. epochs (b) training accuracy vs. epochs (c) test loss vs. epochs (d) test accuracy vs. epochs*

Increasing the layer depth from 16 to 19 and followed by that the increase in the trainable parameters, the Gravity optimizer had more capacity to train the test dataset and show better results than results showed for VGG16 architecture. In all four sections, a major difference can be observed in the behavior of the Gravity optimizer; more speed in reducing loss, achieving less loss, faster speed in increasing accuracy, further reducing validation loss compared to others, faster increasing for validation accuracy, and achieving more validation accuracy. The results were obtained without the use of any overfitting prevention techniques. The values of loss, accuracy, validation loss, and validation accuracy for the training of all five datasets mentioned in Table 3 on two VGGNet architectures, VGG16 and VGG19, using gravity optimizer and two common standard optimization techniques, Adam and RMSProp, are available in details in the Gravity optimizer GitHub repository.

## 5. Conclusion

In this article, we introduced the Gravity optimizer with the idea of treating loss value as height and assuming an acceleration to reducing it. We also used an alternative to moving average. The implementation of the Gravity optimizer is also easily done with the help of Python and TensorFlow, which is available in the Gravity GitHub repository. Gravity optimizer showed acceptable results in performed experiments. In future works, studying on choosing more effective Max-Step Grad, implementing the Gravity in other frameworks like PyTorch, experimenting with larger datasets like ImageNet and larger and more novel architectures and techniques, and using the Gravity for problems other than computer vision such as sequence models and NLP is suggested.

**Biographies:**

*Dariush Bahrami* was born in 1993 in Iran. He completed his undergraduate studies in Fluid Mechanics in 2018. He is currently a student of micro electromechanics systems engineering at the University of Tehran. Most of his previous research has involved computer simulations of physical phenomena. Currently, due to his interest in artificial intelligence and especially artificial neural networks, he is focusing on research in this field.

*Sadegh Pouriyan Zadeh* was born in Kermanshah in 1995. He received his B.Sc. in Electronics Engineering from the Razi University of Kermanshah in 2018. Now he is studying Micro-Electro-Mechanical Systems Engineering in University of Tehran. Although working on gas sensors by nanofibers in his master's thesis, his research interests are in the field of artificial intelligence and its branches.